\documentclass[10pt,twocolumn,letterpaper]{article}
\usepackage{stfloats}
\usepackage{iccv} 
\usepackage{times} 
\usepackage{epsfig} 
\usepackage{graphicx} 
\usepackage{amsmath} 
\usepackage{amssymb} 
\usepackage{subfigure} 
\usepackage{algorithm} 
\usepackage{algorithmicx} 
\usepackage{booktabs} 
\usepackage{threeparttable} 
\usepackage{multirow}
\usepackage{color}
\usepackage{capt-of,etoolbox}

\usepackage[pagebackref=true,breaklinks=true,letterpaper=true,colorlinks,bookmarks=false]{hyperref}

\iccvfinalcopy 

\makeatletter
\apptocmd\@maketitle{{\myfigure{}\par}}{}{}
\makeatother

\begin{document}

\newcommand\myfigure{
\centering
\includegraphics[width=1.01\linewidth]{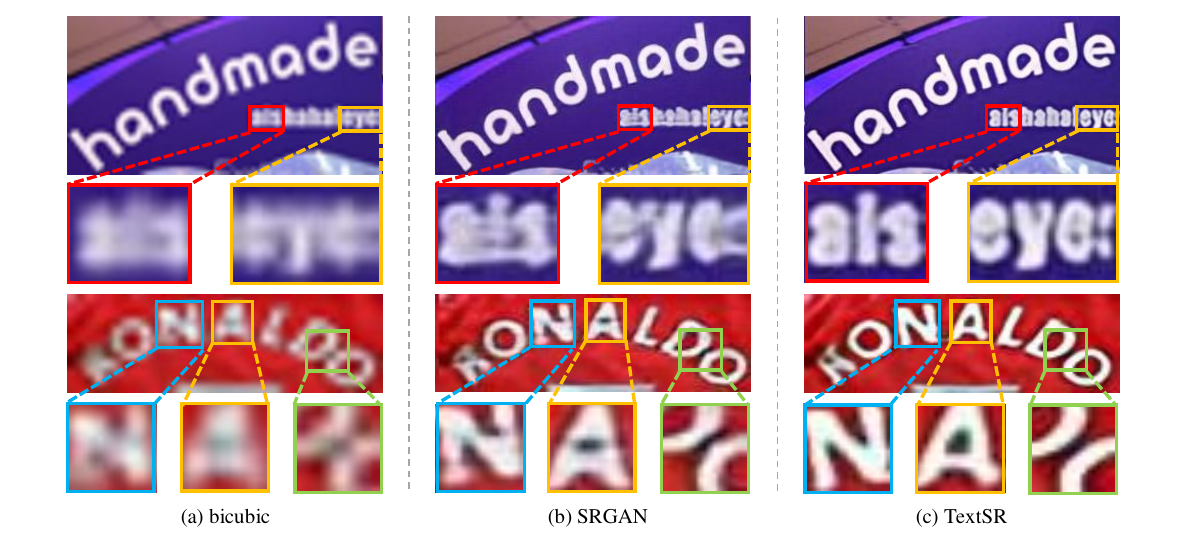}
\captionof{figure}{Super-resolution results from bicubic and SRGAN~\cite{srgan} and our TextSR. The groundtruth of first image is ``handmade, aishahaleye''  and second image is ``RONALDO''. Visual results show that TextSR can successfully restore the content details of texts while bicubic and SRGAN fail to reconstruct the text images.}
\vspace{5mm}
\label{fig:teaser}
}

\title{TextSR: Content-Aware Text Super-Resolution Guided by Recognition}
\author{
{\large
Wenjia Wang$^{1*}$, ~ Enze Xie$^{2}$\thanks{ indicates equal contribution.}, ~ Peize Sun$^{3}$, ~ Wenhai Wang$^{4}$, ~ Chunhua Shen$^{5}$, 
} 
\\[.21cm]
{\large
${^1}$TongJi University ~~~~${^2}$The University of Hong Kong}\\
{\large
${^3}$Xi'an Jiaotong University ~~~~${^4}$Nanjing University ~~~${^5}$The University of Adelaide}
}
\maketitle

\begin{abstract}

Scene text recognition has witnessed rapid development with the advance of convolutional neural networks. Nonetheless, most of the previous methods may not work well in recognizing  text with low resolution which is often seen in natural scene images. An intuitive solution is to introduce super-resolution techniques as pre-processing.
However, conventional super-resolution methods in the literature mainly focus on reconstructing the detailed texture of natural images, 
which typically do not work well for text due to the unique characteristics of text. To tackle these problems, in this work, we propose a content-aware text super-resolution network to generate the information desired for text recognition. In particular,we design an end-to-end network that can perform super-resolution and text recognition simultaneously.
Different from previous super-resolution methods, we use the loss of text recognition as the Text Perceptual Loss to guide the training of the super-resolution network, and thus it 
pays more attention to the text content, rather than the irrelevant background area. Extensive experiments on several challenging benchmarks demonstrate the effectiveness of our proposed method in restoring a sharp high-resolution image from a small blurred one, and show that the recognition performance clearly boosts up the performance of text recognizer. To our knowledge, this is the first work focusing on text super-resolution. 
Code 
is available at:
{\color{red}\url{https://github.com/xieenze/TextSR}}. 
\end{abstract}

\section{Introduction}
Scene text recognition is a fundamental and important task in computer vision, since it is usually a key step towards many downstream text-related applications, including document retrieval, card recognition, and many other Natural Language Processing~(NLP) related applications. Text recognition has achieved remarkable success due to the development of Convolutional Neural Network~(CNN) and text detection~\cite{spcnet,psenet,pan}.

Many accurate and efficient methods have been proposed for most constrained scenarios (e.g., text in scanned copies or network images).
Recent works focus on texts in natural scenes, 
which is much more challenging due to the high diversity of texts in blur, orientation, shape and low-resolution. A thorough survey on recent advantages of text recognition can be found in ~\cite{long2018scene}.

Modern
text recognizers have achieved impressive results on
clear text images.
However, their performances drop sharply when recognizing  blurred text caused by low resolution or camera shake.
The main difficulty to recognize  blurred text 
is the lack of detailed information about them.
Super-resolution~\cite{srgan} is a 
plausible 
method 
to tackle this problem.
However, traditional super-resolution methods aim to reconstruct the detailed texture of natural images, which is not applicable to the blurred text.
Compared to 
the texture of natural images, scenes texts are of arbitrary poses, illumination and blur, super-resolution on scene text images is much more challenging.
Therefore, \emph{we need a content-aware text super-resolution network to generate clear, sharp and identifiable text images for recognition}.

\begin{figure*}[t]
\centering
\includegraphics[width=1.01\textwidth]{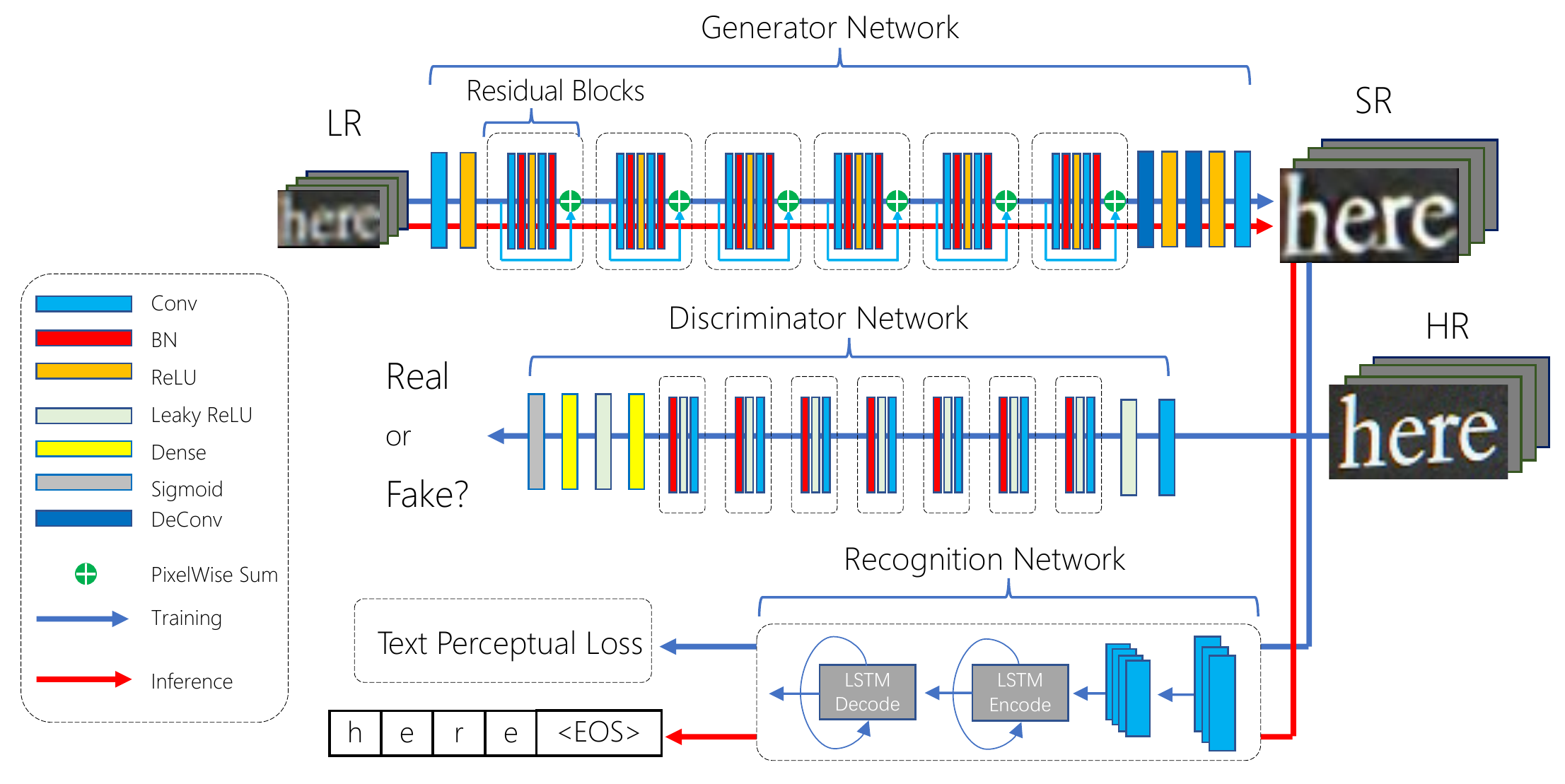}
\caption{The overall architecture of TextSR. It contains three components. The first line is the generator for super-resolution. The second line is the discriminator to distinguish whether the image is high-resolution or generated super-resolution. The last line is the recognizer and TPL to help optimize the generator to produce content-aware text images.}
\label{fig:pipeline}
\end{figure*}

To deal with these 
problems, 
we propose a content-aware text super-resolution network (TextSR), which combines a super-resolution network and a text recognition network.
TextSR is an end-to-end network, in which 
the results of the text recognition network can feed back to guide the training of the super-resolution network.
Under the guidance of the text recognition network, \emph{the super-resolution network would focus on refining the text region, and thus generate clear, sharp and identifiable text images}.
As shown in Fig.~\ref{fig:pipeline}, there are three main components in our network: generator, discriminator and text recognizer. In the generator, a super-resolution network is used to up-sample the small blurred text to a fine scale for recognition.
Compared with
bilinear and bicubic interpolation, the generator can partly reduce artifacts and improve the quality of up-sampled images with a large up-scaling factors.
In the discriminator, a classification network is applied to distinguish the high-resolution image and generated super-resolution image for adversarial training.
Nevertheless, even with such a sophisticated generator and discriminator, up-sampled images are unsatisfactory, usually blurring and lacking fine details, due to the lack of the text content-aware capability.
Therefore, 
we introduce a novel Text Perceptual Loss (TPL) to make the generator produce identifiable and clear text images. 
The TPL is provided by the text recognizer to 
guide the generator to produce clear texts for easier recognition.


The contributions of this work are therefore three-fold: 
\begin{itemize}
\itemsep -0.1cm
    \item 

We introduce a super-resolution method to facilitate scene text recognition, especially for small blurred text. 
\item

We propose a novel Text Perceptual Loss to make the generator be aware of content of text and produce recognition-friendly information. 

\item 
We demonstrate the effectiveness of our proposed method on several challenging public benchmarks and outperforms the strong baseline approaches.
\end{itemize}

\section{Related Work}

\textbf{Super-Resolution} Super-resolution aims to output a plausible  high-resolution image that is consistent with a given low-resolution image. Classical approaches, such as bilinear, bicubic or designed filtering, leverage the insight that neighbouring pixels usually exhibit  similar colours and generate the output by interpolating between the colours of neighbouring pixels according to a predefined formula. Data-driven approaches make use of training data to assist with generating high-resolution predictions and directly copy pixels from the most similar patches in the training dataset~\cite{freedman2011image}. In contrast,  optimization-based approaches, formulate the problem as a sparse recovery problem, where the dictionary is learned from data~\cite{gu2015convolutional}. In the deep learning era, super-resolution 
is simply treated as a regression problem, where the input is the low-resolution image, and the target output is the high-resolution image~\cite{dong2015image}. A deep neural net is trained on the input and target output pairs to minimize some distance metric between the prediction and the ground truth. Subsequently, it was shown that enabling the network to learn the up-scaling filters directly can further increase performance both in terms of accuracy and speed~\cite{dong2016accelerating}. 

\textbf{Generative Adversarial Network} Generative adversarial network (GAN)~\cite{gan} introduces the adversarial loss to generate realistic-appearance images from random noises and achieves impressive results in many generating tasks, such as image generation, image editing, representation learning, image annotation, character transferring~\cite{isola2017image}. GAN is firstly applied to super-resolution by~\cite{srgan} (SRGAN) and obtains promising results on natural images. However, the high resolution images directly generated by SRGAN lack fine details desired for the recognition task. Therefore, we put forward content-aware super-resolution pipeline to recover the information friendly for text recognition.

\textbf{Text Recognition}
The task of text recognition is to recognize character sequences from the cropped word image patches. With the rapid development in deep learning, a large number of effective frameworks have emerged in text recognition. Early work~\cite{jaderberg2014} adopts a bottom-up fashion which detects individual characters firstly and integrates them into a word, or a top-down manner, which treats the word image patch as a whole and  recognizes it as a multi-class image classification problem.  Considering  that  scene text generally appears as a character sequence, following works regard it as a sequence recognition problem and employs Recurrent Neural Network (RNNs) to model the sequential features. For instance, the Connectionist Temporal Classification(CTC)~\cite{ctc} loss is often combined with the RNN outputs for calculating the conditional probability between the predicted sequences and the target~\cite{liu2016star,liu2018squeezedtext}.  Recently, an increasing number of recognition approaches based on the attention mechanism have achieved significant improvements~\cite{cheng2017focusing,luo2019moran}. Among them, ASTER~\cite{aster} rectified oriented or curved text based on Spatial Transformer Network(STN)~\cite{stn} and then performed recognition using an attentional sequence-to-sequence  model. In this work, we choose ASTER as our baseline.

\section{Method}

In this section, we present our proposed method in detail. First, we give a brief description on the SRGAN. Then, the 
overall 
architecture of our method TextSR is presented, as shown in Fig.~\ref{fig:pipeline} and our novel Text Perceptual Loss.

\subsection{SRGAN}
The goal of super-resolution is to train a function that estimates for a given low-resolution input image its corresponding high-resolution counterpart. To achieve this, a generator network $G$, parametrized by $\theta_G$, is optimized by super-resolution specific loss function $l^{SR}$ on training images $I_n^{HR}$ and corresponding $I_n^{LR} (n=1,..,N)$:
\begin{equation}
  \mathop{\arg} \mathop{\min}_{\theta_G}
  \frac{1}{N} \sum_{n=1}^{N} l^{SR} (G_{\theta_G}(I_n^{LR}), I_n^{HR})
\end{equation}

SRGAN defines $l^{SR}$ as the weighted sum of the content loss and the adversarial loss. Instead of 
employing
the pixel-wise loss, it uses the perceptual loss~\cite{johnson2016perceptual} as the content loss.  The adversarial loss is implemented by a discriminator network $D$, parametrized by $\theta_D$, which is optimized in an alternating manner along with $G$ to solve the adversarial min-max problem:
\begin{equation}
\begin{aligned}
    \mathop{\arg} \mathop{\min}_{\theta_G} \mathop{\max}_{\theta_D} ~
    &\mathbb{E}_{I^{HR} \sim p_{data}(I^{HR})}[\log~D_{\theta_D}(I^{HR})] + \\ 
    &\mathbb{E}_{I^{LR} \sim p_G(I^{LR})}[\log~(1-D_{\theta_D}(G_{\theta_G}(I^{LR})))]
\end{aligned}
\end{equation}

The traditional approach of leveraging SRGAN to help the task of text recognition is to separately train a generator that transforms the low-resolution image to high-resolution under the guidance of the adversarial loss. However, such an conventional generator may focus on reconstructing the detailed texture of natural images, which is not applicable to the text content. A more effective text super-resolution network needs a content-aware generator to produce clear, sharp and identifiable text images for recognition, rather than more details of the irrelevant background area. Therefore, we propose our method, TextSR.  



\subsection{Network Architecture}
Our generator network up-samples a low-resolution image and outputs a super-resolution image. Second, a discriminator is used to distinguish whether the images  belong to SR or HR. Furthermore, we add an additional text recognition branch to guide the generator produce content-aware images.

\textbf{Generator network.}
As shown in Fig.~\ref{fig:pipeline}, we adopt a deep CNN architecture which has shown effectiveness for image super-resolution in~\cite{srgan}.
There are two deconvolutional layers in the network, and each layer consists of learned kernels which perform up-sampling a low-resolution image to a 2$\times$ super-resolution image. We use the batch normalization and rectified linear unit (ReLU) activation after each convolutional layer except the last layer.
The generator network can up-sample a low-resolution
image and outputs a 4$\times$ super-resolution image.

\textbf{Discriminator network.}
We employ the same architecture as in~\cite{srgan} for our backbone network in the discriminator, as shown in Fig.~\ref{fig:pipeline}.
The input is the super-resolution image or HR image, and the output is the probability of the input being an HR image. 

\textbf{Recognition Network.}
To show the effectiveness of our method, we adopt ASTER \cite{aster} to be our base recognition network.
ASTER is a state-of-the-art text recognizer composed of a text rectification network and a text recognition network. 
The text rectification network is able to rectify the character arrangement in the input image by
using Thin-Plate-Spline~\cite{tps}, a non-rigid deformation transformer to rearrange the irregular text into horizontal one. 

Based on the rectified text image, 
the text recognition network directly predicts the sequence of characters through sequence-to-sequence translation. 
The text recognition network consists of two parts: encoder and decoder. The encoder is used to extract the feature of the rectified text image. 
It consists of residual blocks~\cite{he2016deep}. 
Following the residual blocks, the feature map is converted into a feature sequence by being split along its row axis. There are two layers of Bidirectional LSTM~(BLSTM)~\cite{graves2008novel} to capture long-range dependencies in both directions. Each one consists of a pair of LSTMs. The outputs of the LSTMs are concatenated and through linearly projected layers before entering the next layer. The decoders are attentional LSTMs to recognizes 94 character classes, including digits, upper-case and lower-case letters, and 32 ASCII punctuation marks. At each time step, it predicts either a character or an end-of-sequence symbol (EOS). The loss of text recognition $l^{TR}$ is defined as:
\begin{equation}
    l^{TR} = - \sum_{t=1}^{T} ( \log ~ p(y_{t}|I^{SR}))
\end{equation}
where $T$ is the length of symbol sequence, $I^{SR}$ is the generated super-resolution image, $y_{1},\cdots, y_{t},\cdots,y_{T}$ is the ground-truth text represented by a character sequence,  $p$ is the corresponding output of decoder.


\subsection{Text Perceptual Loss}
\label{TPL}
Training generator network by only the adversarial loss from discriminator leads to the generator focusing on reconstructing the detailed texture of images without understanding the content in the text image. 
It works on natural images, but it is not practical on text images because content is more important than texture in text images.
Here, we introduce Text Perceptual Loss~(TPL) inspired by the perceptual loss, which is widely used in super-resolution and other low-level vision tasks. 
The perceptual loss uses a pre-trained VGG~\cite{vgg} network and calculate the similarity of the feature map of super-resolution images and original images. Perceptual loss can make network understand the general content of the image since VGG is pre-trained on ImageNet, which contains 1000 kinds of objects. 

To this end, we carefully design TPL by back-propagating the loss of text recognition $l^{TR}$ into the training of the generator network. Specifically, the super-resolution images produced by the generator is directly put into text recognizer. As a result, the generator is trained to minimize $l^{TR}$ on training images $I_n$ and corresponding text characters $y_n (n=1,...,N):$

\begin{equation}
  \mathop{\arg} \mathop{\min}_{\theta_G}
  \frac{1}{N} \sum_{n=1}^{N} 
  l^{TR} (G_{\theta_G}(I_n^{LR}), y_n)
\end{equation}

TPL helps the generator produce more content-realistic images, which are more friendly for the text recognizer.
There are three approaches 
to supervise the generator by TPL. Details are shown as belows:  

\begin{itemize}
    \item End-to-end training ASTER and generator simultaneously from random initialization.
     \item Training ASTER first and then end-to-end training ASTER and generator simultaneously.
    \item Training ASTER first and then training generator while freezing the parameters of ASTER.
   
\end{itemize}

 Our experiments find that they result in very similar performance so we only report the results of third approach.

During training, we prepare the LR-HR image pairs and train three sub-nets shown at the blue line in Fig.~\ref{fig:pipeline}. 
During inference, we select those images which size are smaller than 128$\times$32, then put them into the generator. Finally these restored images are recognized by ASTER. The inference step is shown at red lines in  Fig.~\ref{fig:pipeline}.


\begin{figure*}[t]
\centering
\includegraphics[width=1.01\textwidth]{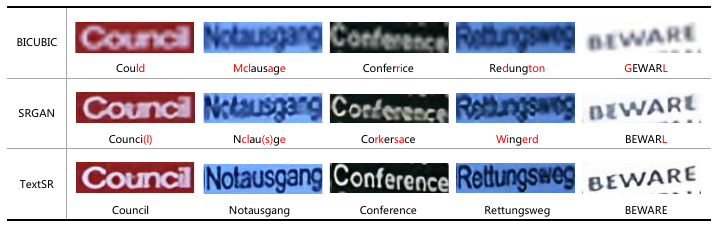}
\caption{Our TextSR Super-resolution results compared with BUCUBIC and SRGAN. It shows that our TextSR generated images can restore the barely recognizable inputs precisely in the detailed textures and shapes, which can remarkably boost up the recognition accuracy. }
\label{fig:exp}
\end{figure*}

\section{Experiments}

We evaluate the proposed method on several challenging public benchmarks and compare its performance with other state-of-the-art methods.

\subsection{Datasets}
The proposed model is trained on a combination of a synthetic dataset and the training set of ICDAR, without fine-tuning on other datasets. The model is tested on 7 standard datasets to evaluate its general recognition performance.

\textbf{Synth90k} is the synthetic text dataset proposed in~\cite{syn90k}. The dataset contains 9 million images. Words are rendered onto natural images with random transformations and effects. All of the images in this dataset are taken for training.

\textbf{SynthText} is the synthetic text dataset proposed in~\cite{syn800k}. It is proposed for text detection. We crop the words using the groundtruth word bounding boxes.

\textbf{IIIT5k-Words} (IIIT5k)~\cite{iiit5k} contains 3000 test images. Each image is associated with a short, 50-word lexicon and a long, 1000-word lexicon.

\textbf{Street View Text} (SVT)~\cite{SVT} is collected from the Google
Street View. The test set contains 647 images of cropped words. 
Each image is associated with a 50-word lexicon.

\textbf{ICDAR 2003} (IC03)~\cite{icdar2003} contains 860 images of cropped word after filtering, discarding words that contain non-alphanumeric characters or have less than three characters. Each image has a 50-word lexicon defined in~\cite{SVT}.

\textbf{ICDAR 2013} (IC13)~\cite{icdar2013} inherits most images from IC03
and extends it with new images. The dataset is filtered by removing words that contain non-alphanumeric characters. The dataset contains 1015 images. No lexicon is provided.

\textbf{ICDAR 2015 Incidental Text} (IC15)~\cite{icdar2015} 
contains a lot of irregular text. Testing images are obtained by cropping the words using the groundtruth word bounding boxes.

\textbf{SVT-Perspective} (SVTP) is proposed in~\cite{svtp} for evaluating the performance of recognizing perspective text. Images in SVTP are picked from the side-view images in Google Street View. The dataset consists of 639 cropped images for testing.

\textbf{CUTE80} (CUTE) is proposed in~\cite{cute} for the curved text. It contains 80 high-resolution images taken in natural scenes. We crop the annotated words and obtain a test set of 288 images.

\subsection{Implementation Details}
During training, we set the trade-off weights of all losses as 1. We use the Adam optimizer with momentum term 0.9. The generator network and discriminator network are trained from scratch and the weights in each layer are initialized with a zero-mean Gaussian distribution with standard deviation 0.02, and biases are initialized with 0. We use the recognizer ASTER from the source code\footnote{\url{https://github.com/ayumiymk/aster.pytorch}}, whose performance is slightly different from the original paper.

The model is trained by batches of 256 examples for 50k iterations with 4 Tesla M40 GPUs. For super-resolution tasks, we use SynthText as our training data and filter the images whose size is smaller than 128$\times$32. In this way we only use 1.29 millions images from SynthText. 
The low-resolution images are 4$\times$ down-sampled  from the original images.
The learning rate is set to 1.0 initially and decayed to 0.1 and 0.01 at step 30k and 40k respectively.
\begin{table}[t]
    \scalebox{0.97}{
    \begin{tabular}{|c|ccccc|} 
    \hline
    Size & 128$\times$32 & 64$\times$16 & 32$\times$8 & 24$\times$6 & 20$\times5$\\
    \hline
    BICUBIC & 90.5 & 89.7 & 63.9 & 30.8 & 16.1   \\
    SRGAN   & 90.6 & 90.1 & 72.7 & 44.9 & 20.0  \\
    TextSR  & \textbf{91.3} & 
    \textbf{90.5} & 
    \textbf{83.1} & 
    \textbf{62.6} & 
    \textbf{42.8} \\
    \hline
    Improve & +0.7 & +0.4 & +10.4 & +17.7 & +22.8\\
    \hline
    \end{tabular}
    }
    \vspace{0.051cm}
	\caption{Text Recognition accuracy without SR, with SRGAN and TextSR  on IC13 dataset. As the input size decreases, the TextSR can impressively boost up the performance on very tiny text images.}
	\label{tab:content-aware}
\end{table}

\begin{table}[!t]
    \scalebox{1}{
    \begin{tabular}{|c|cccc|} 
    \hline
    \textbf{IC13} & BIUCBIC & SRGAN & TextSR & Improve \\
    \hline
     PSNR & 18.65  & 23.77  & \textbf{25.28} & +1.51 \\
     SSIM & 0.7304  & 0.8505& \textbf{0.8823} & +0.0318\\
    \hline
    \hline
    \textbf{IC15} &&&&\\
    \hline
    PSNR & 19.56 & 24.40  & \textbf{25.13}  & +0.73\\
     SSIM & 0.7738  &0.8884& \textbf{0.8993} & +0.0109\\
    \hline
    \hline
    \textbf{IC03} &&&&\\
    \hline
    PSNR & 18.61 & 23.42  & \textbf{24.35}& +0.93  \\
     SSIM & 0.7299  &0.8490 & \textbf{0.8765}& +0.0275\\
     \hline
     \hline
    \textbf{SVT} &&&&\\
    \hline
    PSNR & 21.51 & 26.39  & \textbf{27.74} & +1.35 \\
    SSIM & 0.7932  &0.8948 & \textbf{0.9140} & +0.0192\\
    \hline
    \hline
    \textbf{SVTP} &&&&\\
    \hline
    PSNR & 21.93 & 25.80 & \textbf{25.83} &  +0.03\\
     SSIM & 0.8063  &0.8866 & \textbf{0.8915}&  +0.0049\\
      \hline
      \hline
    \textbf{IIIT5K} &&&&\\
    \hline
    PSNR & 16.48 & 21.01  & \textbf{21.39} &  +0.38\\
     SSIM & 0.7361  &0.8480 & \textbf{0.8648}&  +0.0168\\
      \hline
      \hline
    \textbf{CUTE} &&&&\\
    \hline
    PSNR & 15.56 & 19.85  & \textbf{21.98} &  +2.13\\
     SSIM & 0.6547  &0.8013 & \textbf{0.8561}&  +0.0548\\
     
    \hline
    \end{tabular}
    }
    \vspace{0.051cm}
	\caption{Comparison of methods: BICUBIC, SRGAN and TextSR. Highest calculated measures (PSNR [dB], SSIM) in bold. It can be found that our TextSR clearly surpasses SRGAN in PSNR and SSIM in all  datasets.}
	\label{tab:sr}
\end{table}

\subsection{Ablation Study}
We first show the importance of content-aware on the super-resolution image quality by comparing with the method of directly using SRGAN~\cite{srgan}. Then, we compare our proposed method with the baseline text recognizer ASTER~\cite{aster} to prove the effectiveness of applying our super-resolution strategy to the task of text recognition. Moreover, we visualize activations heatmaps of the generator to show the response on text areas.    
\begin{figure*}[!t]
\centering
\includegraphics[width=0.99\textwidth]{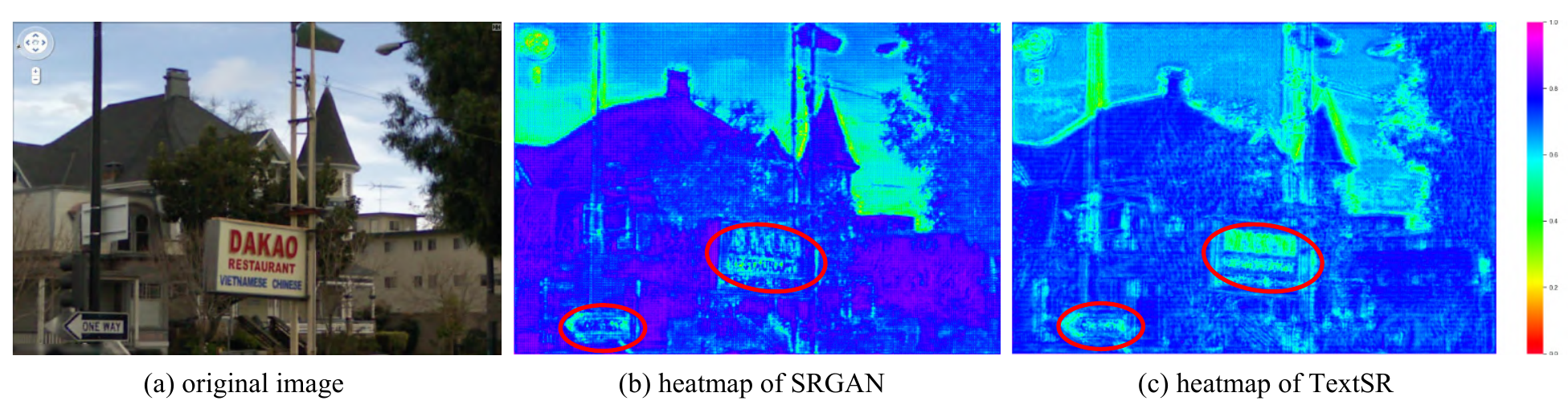}
\caption{The intermediate heatmap visualization of SRGAN and TextSR on the street-view image. It can be found that our TextSR have higher respond than SRGAN in text areas. }
\label{fig:exp3}
\end{figure*}

\begin{figure*}[!t]
    \centering
    \includegraphics[width=1.399\columnwidth]{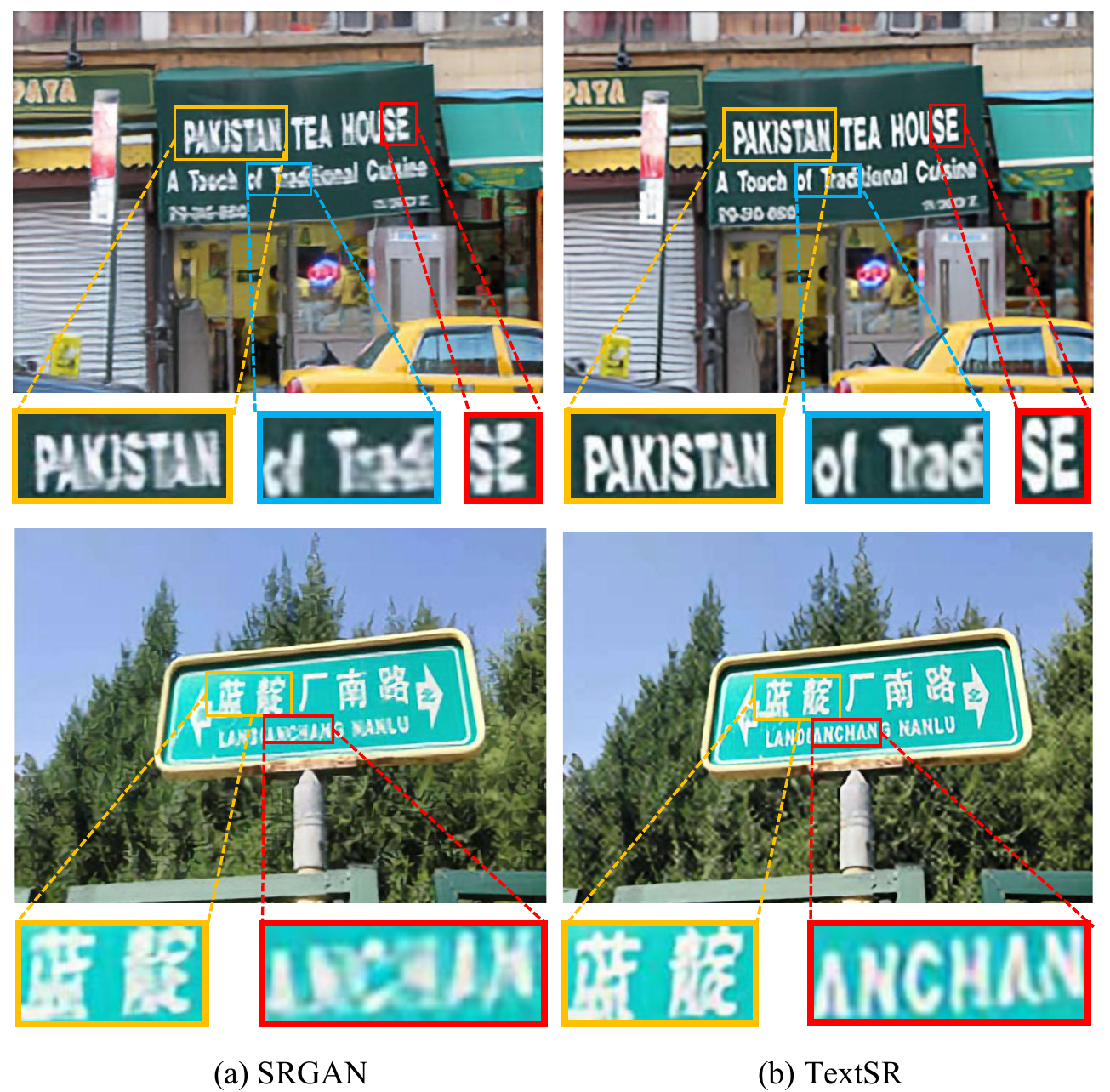}
    \caption{Super-resolution images on scene text detection images without fine-tuning our TextSR. The first image is from SVT detection dataset and the second image is captured by us on the street. It can be easily found that our TextSR has strong ability to restore English texts. But we fail to restore Chinese because we have not trained with Chinese data.}
\label{fig:exp4}
\end{figure*}

\textbf{Effect on super-resolution.} To demonstrate the 
impact
on the super-resolution image quality, we 
carry out 
experiments on 7 datasets to evaluate the SRGAN and TextSR. Details are shown in Table~\ref{tab:sr}. It can be found that our TextSR clearly surpasses SRGAN in PSNR and SSIM in all datasets. Moreover, seen in Fig.~\ref{fig:exp}, our TextSR can indeed generate sharper high-resolution text images than SRGAN. We explain that it is due to the supervision of TPL, so that the generator indeed can understand the content in the text images.

\textbf{Effect on text recognition.} To demonstrate the effect of content-aware TPL, we set up three experiments as shown in Table~\ref{tab:content-aware}.
We down-sample the original images to five different scale and then use three ways to up-sample the images. The first is baseline without super-resolution, the second is with SRGAN. The last is equipped with our content-aware TPL.  Their performance is evaluated on ICDAR2013 test dataset. It can be seen that although SRGAN can partly improve the performance when compared to BICUBIC, our TextSR can greatly boost up the performance in very tiny images. Specially, for images with 20$\times$5, TextSR can improve surprisingly 22.8\% from SRGAN.

Moreover, we show the intermediate activations heatmap. The activations is the penultimate layer of the generator. To obtain a more comprehensive comparison of response on text areas and irrelevant backgrounds, we choose a street-view picture to input the generator and visualize its activation heatmap, shown in Fig.~\ref{fig:exp3}. It can be found that with the guidance of TPL, TextSR can have stronger respond on text area than SRGAN, which explains the reason of TextSR generating more clear, sharp and identifiable text images.

\subsection{Comparison with State-of-the-Art}
We compare the performance of our model with other state-of-the-art models. Some datasets provide lexicons for constraining recognition outputs. When a lexicon is given, we simply replace the predicted word with the nearest lexicon word under the metric of edit distance. 

\begin{table*}[!t]
\centering
\scalebox{.91}{
        \begin{tabular}{|p{4cm}|c|c|cc|c|c|c|c|c|}
	\hline
	\multirow{2}{*}{Method} &
	\multirow{2}{*}{ConvNet,~Data} & {IIIT5K} &
	\multicolumn{2}{c|}{SVT}
	& {IC03} & {IC13} & {IC15} & {SVTP} & {CUTE}\\
	\cline{3-10} & &0& 50 &0& 0 & 0 & 0 &0 &0 \\
	\hline
	Wang, Babenko, and Belongie 2011b\cite{wang2011end} & - & - & 57.0 & - &-  &-  &-  &- &- \\
	\hline
	Mishra, Alahari, and Jawahar 2012b\cite{mishra2012top} & - & - & 73.2 & - & - & - & - & -&-\\
	\hline
	Wang et al. 2012 \cite{wang2012end} & - & - & 70.0 & - & - & - & - &&\\
	\hline
	Bissacco et al. 2013 \cite{bissacco2013photoocr} & - & - & - & - & - & 87.6 & - & - & - \\
	\hline
	Almazan et al. 2014 \cite{almazan2014word} & - & - & 89.2 & - & - & - & - &&\\
	\hline
	Yao et al. 2014 \cite{yao2014strokelets} & - & - & 75.9 & - & - & - & - &&\\
	\hline
	Rodriguez-Serrano, Gordo, and Perronnin 2015 \cite{rodriguez2015label} & - & - & 70.0 & - & - & - & - &&\\
	\hline
	Jaderberg, Vedaldi, and Zisserman 2014 \cite{jaderberg2014} & - & - & 86.1 & - & - & - & - &&\\
	\hline
	Su and Lu 2014 \cite{su2014accurate} & - & - & 83.0 & - & - & - & - &&\\
	\hline
	Gordo 2015 \cite{gordo2015supervised} & - & - & 91.8 & - & - & - & - & - & - \\
	\hline
    Jaderberg et al. 2016 \cite{jaderberg2016reading} & VGG,90k & - & 95.4 & 80.7 & 93.1 & 90.8 & - & - & - \\
	\hline
	Jaderberg et al. 2015a \cite{jaderberg2015deep} & VGG,90k & - & 93.2 & 71.7 & 89.6 & 81.8 & - & - & - \\
	\hline
	Shi, Bai, and Yao 2016 \cite{shi2016end} & VGG,90k & 81.2 & 97.5 & 82.7 & 91.9 & 89.6 & - & - & - \\
	\hline
	Shi et al. 2016 \cite{shi2016robust} & VGG,90k & 81.9 & 95.5 & 81.9 & 90.1 & 88.6 & - &  71.8 & 59.2\\
	\hline
	Lee and Osindero 2016 \cite{lee2016recursive} & VGG,90k & 78.4 & 96.3 & 80.7 & 88.7 & 90.0 & - & - & -\\
	\hline
	Yang et al. 2017 \cite{yang2017learning} & VGG,Private & - & 95.2 & - & - & - & - & 75.8 & 69.3 \\
	\hline
	Cheng et al. 2017 \cite{cheng2017focusing} & ResNet,(90k, ST) & 87.4 & 97.1 & 85.9 & 94.2 & 93.3 & 70.6 & - & - \\
	\hline
	ASTER~\cite{aster} & ResNet,(90k, ST) & 93.4 & 99.2 & 93.6 & 94.5 & 91.8 & 76.1 & 78.5 & 79.5 \\
	\hline
	\hline
	ASTER~(\textit{ReIm}) & ResNet,(90k, ST) & 92.4 & 97.5 & 86.1 & 93.1 & 90.5 & 74.8 & 76.7 & 78.8 \\
	\hline
	ASTER~(\textit{ReIm}) + TextSR & ResNet,(90k, ST) &92.5  & 98.0 &87.2  &93.2  &91.3  & 75.6 &77.4 &78.9 \\
	\hline
	ASTER~(\textit{ReIm}) & ResNet,(90k, ST, FT) & 95.6 & 99.4 & 95.1 & 96.5 &92.1&77.5 &86.7 & 85.4 \\
	\hline
	ASTER~(\textit{ReIm}) + TextSR & ResNet,(90k, ST, FT) &95.6  & 99.4 &95.1  &96.5  &92.4 &79.0 &87.1 &87.2 \\
	\hline
	\end{tabular}
    }
    \vspace{0.051cm}
	\caption{Performance across a number of methods and datasets. ``ReIm'' is our re-implemented ASTER. 50 are lexicons. 0 means no lexicon. ST and 90k means SynthText and Synth90k datasets. Private means using private training data. FT means fine-tuning on real datasets.}
	\label{tab:cross}
\end{table*}

Table~\ref{tab:cross} compares the recognition accuracy across a number of methods. Our re-implementation of ASTER text recognizer is slightly different from the original paper. Equipped with TextSR, it achieves better performance in all of datasets, showing the effectiveness of our proposed method.

To further test whether TextSR can improve performance of the stronger text recognizer, we fine-tune ASTER on real datasets and obtain the more advanced model. Surprisingly, TextSR still boosts up its performance. Particularly, the improvement on IC15 reaches 1.5\%. We claim it arises from that IC15 contains a large amount of small text images and TextSR greatly solves the problem.


\subsection{Extension on text detection images.}

We extend our TextSR on text detection images. It is more challenging because these images contain more irrelative background. The visualization results are shown in Fig.~\ref{fig:exp4}. In the first line we find that our TextSR results can automatically super-resolution text areas with out any detection.

To further evaluate the robustness of TextSR, we randomly choose a image in the street captured by our mobile phone. It is interesting that our TextSR can successfully restore English area but fail to restore Chinese area. We argue that it is obviously because we do not train the model with Chinese data. If we add Chinese data in training process, the restore of Chinese text can be immediately improved.

\section{Conclusion}

This work addresses the problem of small text recognition with content-aware super-resolution. 
To our knowledge, this is the first work attempting to solve small texts with super-resolution methods.
We elaborately add a novel Text Perceptual Loss (TPL) to help the generator restore the content in text images. Compare to standard super-resolution methods, the proposed method pays more attention on generating information of the text itself, rather than texture of background area. It shows superior performance on various text recognition benchmarks. In the future, we will focus on reconstruct text from incomplete text images.

\appendix 

\section{Appendix}
\subsection{More Image Restoration On Recognition Datasets and Detection Datasets}
In this section, we show more test examples produced by TextSR on IC03, IC13, IC15 and CUTE80 in Fig.~1 to Fig.~4. Fig.~1 shows the super-resolution ability on recognition datasets and Fig.~2 to Fig.~4 shows the visual results on detection datasets.

From these results, we can find that the proposed TextSR have the following abilities. (1) Restore the content of text images correctly. (2) Automatically restore text areas in the full images. Thanks to Text Perceptual Loss, the generator can understand the content of images and produce content-realistic images.

\subsection{Failure Cases Analysis}
We analyse two kinds of failure cases in Fig.~\ref{fig:failcase}.
The first type of our failure cases are caused by our generator network. 
Some of the characters in input images are too blur to distinguish from other similar characters. TextSR can dramatically restore the characters to a  similar one completely. This is shown in (a),(b),(c), which separately restore the `C' to `G', `c' to `e', `l' to `i'. In (d), the blurred character `A' is restored to a reversed `K' because the network has learn the reversed shape of the characters with the mechanism of BLSTM. 

The second type of failure cases are caused by the recognizer network. The Generator restored the characters to a fairly good extent due to its excellent ability. But these characters themselves are too confusing so the recognition network failed to recognize them. As is shown in (e) and (f), the characters are of very rare font or irregular shape.

\begin{figure*}[b]
\centering
\includegraphics[width=1.01\textwidth]{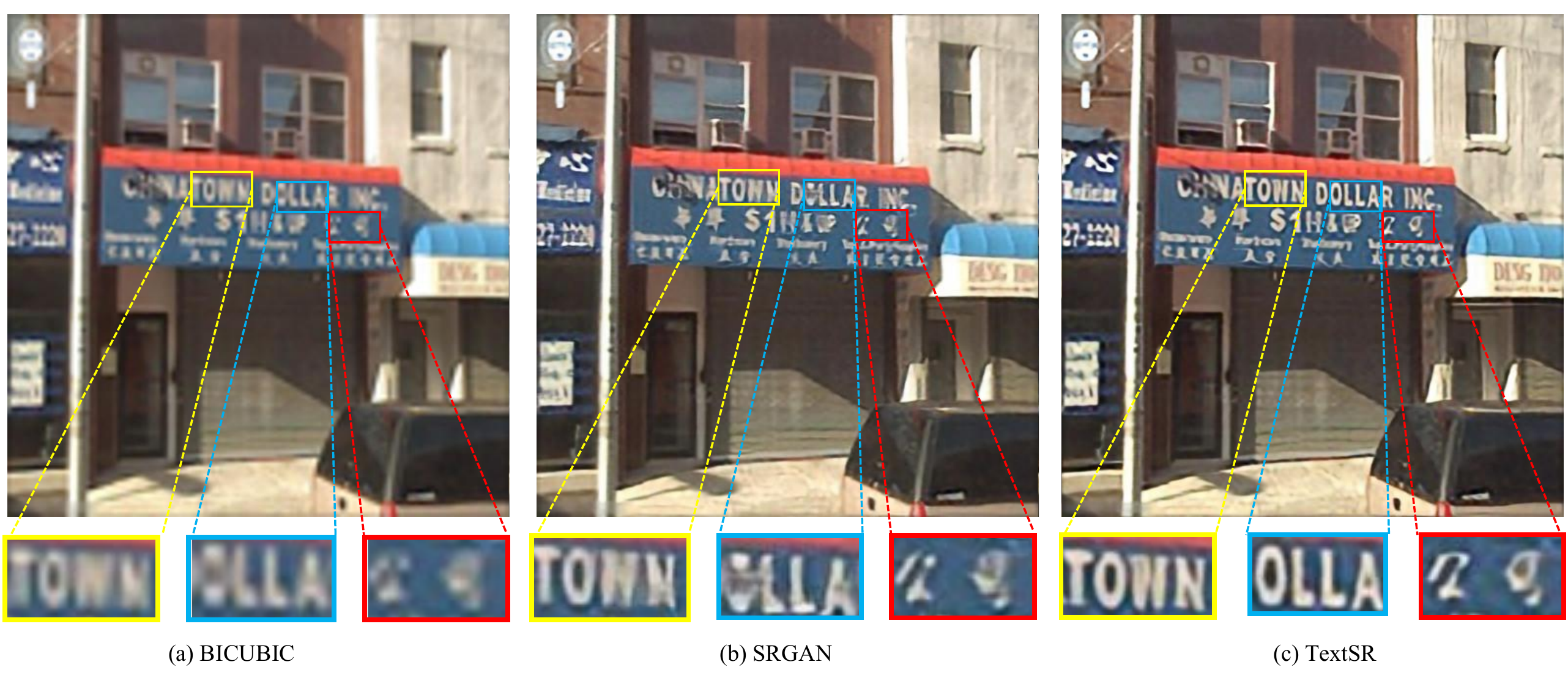}
\caption{Figure from SVT dataset. This picture contains both English and Chinese words.
 }
\label{fig:det3}
\end{figure*}

\begin{figure*}[b]
\centering
\includegraphics[width=0.8\textwidth]{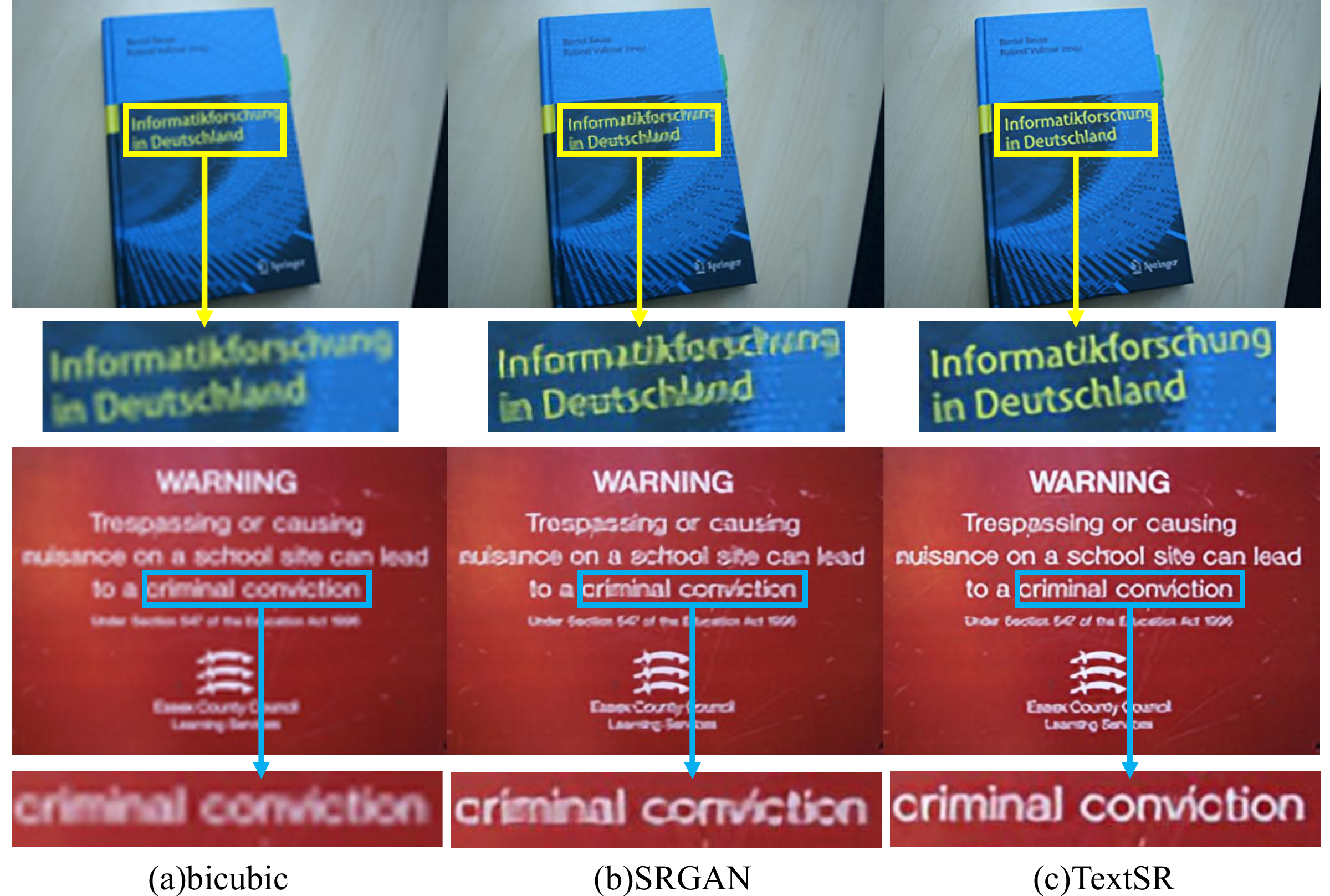}
\caption{Pictures from ICDAR2013 word localization. The groundtruths of the boxed regions are separately `Informatikforschung in Deutschland' and `criminal conviction.'
 }
\label{fig:det}
\end{figure*}

\begin{figure*}[b]
\centering
\includegraphics[width=.8101\textwidth]{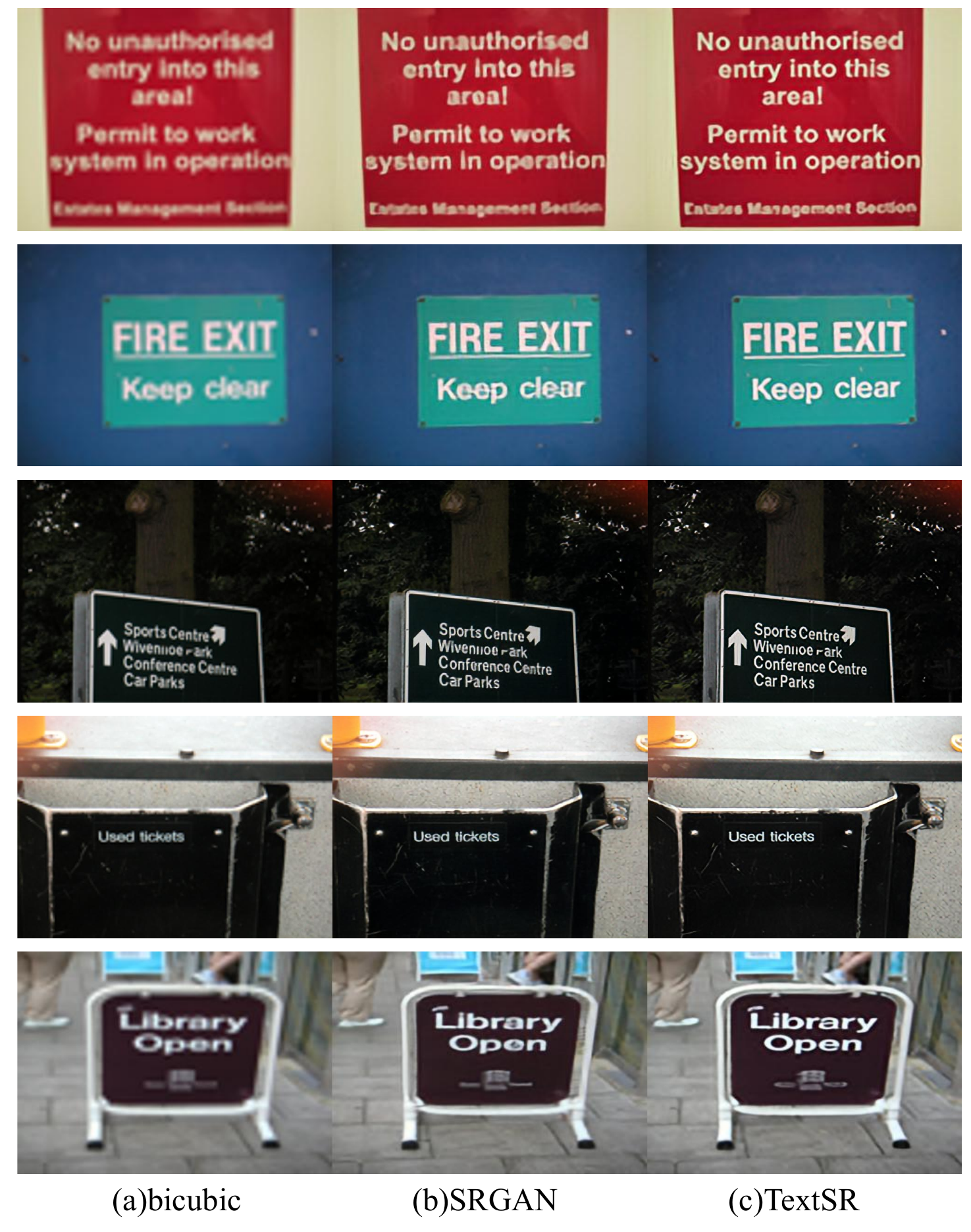}
\caption{Images from ICDAR2013 word localization. The characters restored by SRGAN are hardly recognizable. While those in TextSR are precisely restored especially in those with arcs like `e', `a', `s' etc.
 }
\label{fig:det2}
\end{figure*}

\begin{figure*}[ht]
\centering
\includegraphics[width=0.801\textwidth]{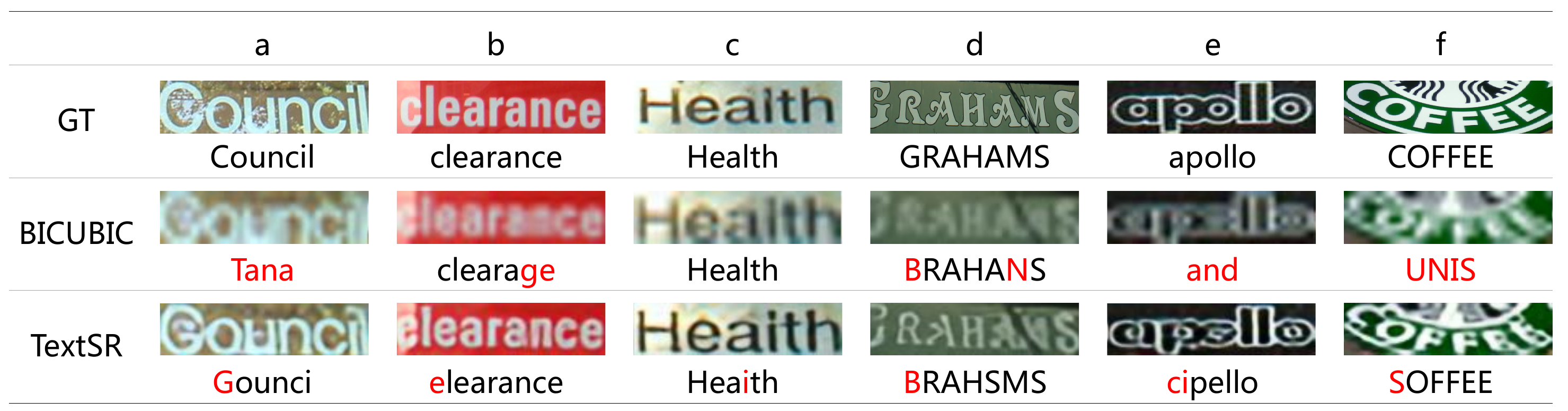}
\caption{Failure Samples.
 }
\label{fig:failcase}
\end{figure*}

\bibliographystyle{ieee}
\bibliography{aaai}

\begin{thebibliography}{10}\itemsep=-1pt

\bibitem{almazan2014word}
J.~Almaz{\'a}n, A.~Gordo, A.~Forn{\'e}s, and E.~Valveny.
\newblock Word spotting and recognition with embedded attributes.
\newblock {\em {IEEE} Trans. Pattern Anal. Mach. Intell.}, 2014.

\bibitem{bissacco2013photoocr}
A.~Bissacco, M.~Cummins, Y.~Netzer, and H.~Neven.
\newblock Photoocr: Reading text in uncontrolled conditions.
\newblock In {\em Proc. IEEE Int. Conf. Comp. Vis.}, 2013.

\bibitem{tps}
F.~L. Bookstein.
\newblock Principal warps: Thin-plate splines and the decomposition of
  deformations.
\newblock {\em {IEEE} Trans. Pattern Anal. Mach. Intell.}, 1989.

\bibitem{cheng2017focusing}
Z.~Cheng, F.~Bai, Y.~Xu, G.~Zheng, S.~Pu, and S.~Zhou.
\newblock Focusing attention: Towards accurate text recognition in natural
  images.
\newblock In {\em Proc. IEEE Int. Conf. Comp. Vis.}, 2017.

\bibitem{dong2015image}
C.~Dong, C.~C. Loy, K.~He, and X.~Tang.
\newblock Image super-resolution using deep convolutional networks.
\newblock {\em {IEEE} Trans. Pattern Anal. Mach. Intell.}, 2015.

\bibitem{dong2016accelerating}
C.~Dong, C.~C. Loy, and X.~Tang.
\newblock Accelerating the super-resolution convolutional neural network.
\newblock In {\em Proc. Eur. Conf. Comp. Vis.}, 2016.

\bibitem{freedman2011image}
G.~Freedman and R.~Fattal.
\newblock Image and video upscaling from local self-examples.
\newblock {\em ACM Trans. Gr.}, 2011.

\bibitem{gan}
I.~Goodfellow, J.~Pouget-Abadie, M.~Mirza, B.~Xu, D.~Warde-Farley, S.~Ozair,
  A.~Courville, and Y.~Bengio.
\newblock Generative adversarial nets.
\newblock In {\em Proc. Adv. Neural Inf. Process. Syst.}, 2014.

\bibitem{gordo2015supervised}
A.~Gordo.
\newblock Supervised mid-level features for word image representation.
\newblock In {\em Proc. IEEE Conf. Comp. Vis. Patt. Recogn.}, 2015.

\bibitem{ctc}
A.~Graves, S.~Fern{\'a}ndez, F.~Gomez, and J.~Schmidhuber.
\newblock Connectionist temporal classification: labelling unsegmented sequence
  data with recurrent neural networks.
\newblock In {\em Proc. Int. Conf. Mach. Learn.}, 2006.

\bibitem{graves2008novel}
A.~Graves, M.~Liwicki, S.~Fern{\'a}ndez, R.~Bertolami, H.~Bunke, and
  J.~Schmidhuber.
\newblock A novel connectionist system for unconstrained handwriting
  recognition.
\newblock {\em {IEEE} Trans. Pattern Anal. Mach. Intell.}, 2008.

\bibitem{gu2015convolutional}
S.~Gu, W.~Zuo, Q.~Xie, D.~Meng, X.~Feng, and L.~Zhang.
\newblock Convolutional sparse coding for image super-resolution.
\newblock In {\em Proc. IEEE Int. Conf. Comp. Vis.}, 2015.

\bibitem{syn800k}
A.~Gupta, A.~Vedaldi, and A.~Zisserman.
\newblock Synthetic data for text localisation in natural images.
\newblock In {\em Proc. IEEE Conf. Comp. Vis. Patt. Recogn.}, 2016.

\bibitem{he2016deep}
K.~He, X.~Zhang, S.~Ren, and J.~Sun.
\newblock Deep residual learning for image recognition.
\newblock In {\em Proc. IEEE Conf. Comp. Vis. Patt. Recogn.}, 2016.

\bibitem{isola2017image}
P.~Isola, J.-Y. Zhu, T.~Zhou, and A.~A. Efros.
\newblock Image-to-image translation with conditional adversarial networks.
\newblock In {\em Proc. IEEE Conf. Comp. Vis. Patt. Recogn.}, 2017.

\bibitem{syn90k}
M.~Jaderberg, K.~Simonyan, A.~Vedaldi, and A.~Zisserman.
\newblock Synthetic data and artificial neural networks for natural scene text
  recognition.
\newblock {\em arXiv preprint arXiv:1406.2227}, 2014.

\bibitem{jaderberg2015deep}
M.~Jaderberg, K.~Simonyan, A.~Vedaldi, and A.~Zisserman.
\newblock Deep structured output learning for unconstrained text recognition.
\newblock {\em Proc. Int. Conf. Learn. Repren.}, 2015.

\bibitem{jaderberg2016reading}
M.~Jaderberg, K.~Simonyan, A.~Vedaldi, and A.~Zisserman.
\newblock Reading text in the wild with convolutional neural networks.
\newblock {\em Int. J. Comp. Vis.}, 2016.

\bibitem{stn}
M.~Jaderberg, K.~Simonyan, A.~Zisserman, et~al.
\newblock Spatial transformer networks.
\newblock In {\em Proc. Adv. Neural Inf. Process. Syst.}, 2015.

\bibitem{jaderberg2014}
M.~Jaderberg, A.~Vedaldi, and A.~Zisserman.
\newblock Deep features for text spotting.
\newblock In {\em Proc. Eur. Conf. Comp. Vis.}, 2014.

\bibitem{johnson2016perceptual}
J.~Johnson, A.~Alahi, and L.~Fei-Fei.
\newblock Perceptual losses for real-time style transfer and super-resolution.
\newblock In {\em Proc. Eur. Conf. Comp. Vis.}, 2016.

\bibitem{icdar2015}
D.~Karatzas, L.~Gomez-Bigorda, A.~Nicolaou, S.~Ghosh, A.~Bagdanov, M.~Iwamura,
  J.~Matas, L.~Neumann, V.~R. Chandrasekhar, S.~Lu, et~al.
\newblock Icdar 2015 competition on robust reading.
\newblock In {\em Proc. IEEE Int. Conf. Doc. Anal. and Recogn.}, 2015.

\bibitem{icdar2013}
D.~Karatzas, F.~Shafait, S.~Uchida, M.~Iwamura, L.~G. i~Bigorda, S.~R. Mestre,
  J.~Mas, D.~F. Mota, J.~A. Almazan, and L.~P. De~Las~Heras.
\newblock Icdar 2013 robust reading competition.
\newblock In {\em Proc. IEEE Int. Conf. Doc. Anal. and Recogn.}, 2013.

\bibitem{srgan}
C.~Ledig, L.~Theis, F.~Husz{\'a}r, J.~Caballero, A.~Cunningham, A.~Acosta,
  A.~Aitken, A.~Tejani, J.~Totz, Z.~Wang, et~al.
\newblock Photo-realistic single image super-resolution using a generative
  adversarial network.
\newblock In {\em Proc. IEEE Conf. Comp. Vis. Patt. Recogn.}, 2017.

\bibitem{lee2016recursive}
C.-Y. Lee and S.~Osindero.
\newblock Recursive recurrent nets with attention modeling for ocr in the wild.
\newblock In {\em Proc. IEEE Conf. Comp. Vis. Patt. Recogn.}, 2016.

\bibitem{liu2016star}
W.~Liu, C.~Chen, K.-Y.~K. Wong, Z.~Su, and J.~Han.
\newblock Star-net: a spatial attention residue network for scene text
  recognition.
\newblock In {\em Proc. Brit. Mach. Vis. Conf.}, 2016.

\bibitem{liu2018squeezedtext}
Z.~Liu, Y.~Li, F.~Ren, W.~L. Goh, and H.~Yu.
\newblock Squeezedtext: A real-time scene text recognition by binary
  convolutional encoder-decoder network.
\newblock In {\em Proc. AAAI Conf. on Arti. Intel.}, 2018.

\bibitem{long2018scene}
S.~Long, X.~He, and C.~Ya.
\newblock Scene text detection and recognition: The deep learning era.
\newblock {\em arXiv preprint arXiv:1811.04256}, 2018.

\bibitem{icdar2003}
S.~M. Lucas, A.~Panaretos, L.~Sosa, A.~Tang, S.~Wong, R.~Young, K.~Ashida,
  H.~Nagai, M.~Okamoto, H.~Yamamoto, et~al.
\newblock {ICDAR} 2003 robust reading competitions: entries, results, and
  future directions.
\newblock {\em Int.\ J.\ Document Analysis and Recognition}, 2005.

\bibitem{luo2019moran}
C.~Luo, L.~Jin, and Z.~Sun.
\newblock Moran: A multi-object rectified attention network for scene text
  recognition.
\newblock {\em Pattern Recognition}, 2019.

\bibitem{iiit5k}
A.~Mishra, K.~Alahari, and C.~Jawahar.
\newblock Top-down and bottom-up cues for scene text recognition.
\newblock In {\em Proc. IEEE Conf. Comp. Vis. Patt. Recogn.}, 2012.

\bibitem{mishra2012top}
A.~Mishra, K.~Alahari, and C.~Jawahar.
\newblock Top-down and bottom-up cues for scene text recognition.
\newblock In {\em Proc. IEEE Conf. Comp. Vis. Patt. Recogn.}, 2012.

\bibitem{svtp}
T.~Quy~Phan, P.~Shivakumara, S.~Tian, and C.~Lim~Tan.
\newblock Recognizing text with perspective distortion in natural scenes.
\newblock In {\em Proc. IEEE Int. Conf. Comp. Vis.}, 2013.

\bibitem{cute}
A.~Risnumawan, P.~Shivakumara, C.~S. Chan, and C.~L. Tan.
\newblock A robust arbitrary text detection system for natural scene images.
\newblock {\em Expert Systems with Applications}, 2014.

\bibitem{rodriguez2015label}
J.~A. Rodriguez-Serrano, A.~Gordo, and F.~Perronnin.
\newblock Label embedding: A frugal baseline for text recognition.
\newblock {\em Int. J. Comp. Vis.}, 2015.

\bibitem{shi2016end}
B.~Shi, X.~Bai, and C.~Yao.
\newblock An end-to-end trainable neural network for image-based sequence
  recognition and its application to scene text recognition.
\newblock {\em {IEEE} Trans. Pattern Anal. Mach. Intell.}, 2016.

\bibitem{shi2016robust}
B.~Shi, X.~Wang, P.~Lyu, C.~Yao, and X.~Bai.
\newblock Robust scene text recognition with automatic rectification.
\newblock In {\em Proc. IEEE Conf. Comp. Vis. Patt. Recogn.}, 2016.

\bibitem{aster}
B.~Shi, M.~Yang, X.~Wang, P.~Lyu, C.~Yao, and X.~Bai.
\newblock Aster: An attentional scene text recognizer with flexible
  rectification.
\newblock {\em {IEEE} Trans. Pattern Anal. Mach. Intell.}, 2018.

\bibitem{vgg}
K.~Simonyan and A.~Zisserman.
\newblock Very deep convolutional networks for large-scale image recognition.
\newblock {\em Proc. Int. Conf. Learn. Repren.}, 2015.

\bibitem{su2014accurate}
B.~Su and S.~Lu.
\newblock Accurate scene text recognition based on recurrent neural network.
\newblock In {\em Proc. Asian Conf. Comp. Vis.}, 2014.

\bibitem{SVT}
K.~Wang, B.~Babenko, and S.~Belongie.
\newblock End-to-end scene text recognition.
\newblock In {\em Proc. IEEE Int. Conf. Comp. Vis.}, 2011.

\bibitem{wang2011end}
K.~Wang, B.~Babenko, and S.~Belongie.
\newblock End-to-end scene text recognition.
\newblock In {\em Proc. IEEE Int. Conf. Comp. Vis.}, 2011.

\bibitem{wang2012end}
T.~Wang, D.~J. Wu, A.~Coates, and A.~Y. Ng.
\newblock End-to-end text recognition with convolutional neural networks.
\newblock In {\em Proc. IEEE Int. Conf. Patt. Recogn.}, 2012.

\bibitem{psenet}
W.~Wang, E.~Xie, X.~Li, W.~Hou, T.~Lu, G.~Yu, and S.~Shao.
\newblock Shape robust text detection with progressive scale expansion network.
\newblock In {\em {IEEE} Conference on Computer Vision and Pattern Recognition,
  {CVPR} 2019, Long Beach, CA, USA, June 16-20, 2019}, pages 9336--9345, 2019.

\bibitem{pan}
W.~Wang, E.~Xie, X.~Song, Y.~Zang, W.~Wang, T.~Lu, G.~Yu, and C.~Shen.
\newblock Efficient and accurate arbitrary-shaped text detection with pixel
  aggregation network.
\newblock {\em CoRR}, abs/1908.05900, 2019.

\bibitem{spcnet}
E.~Xie, Y.~Zang, S.~Shao, G.~Yu, C.~Yao, and G.~Li.
\newblock Scene text detection with supervised pyramid context network.
\newblock In {\em Proceedings of the AAAI Conference on Artificial
  Intelligence}, volume~33, pages 9038--9045, 2019.

\bibitem{yang2017learning}
X.~Yang, D.~He, Z.~Zhou, D.~Kifer, and C.~L. Giles.
\newblock Learning to read irregular text with attention mechanisms.
\newblock In {\em Proc.\ Int. Joint Conf. Aritificial Intelligence}, 2017.

\bibitem{yao2014strokelets}
C.~Yao, X.~Bai, B.~Shi, and W.~Liu.
\newblock Strokelets: A learned multi-scale representation for scene text
  recognition.
\newblock In {\em Proc. IEEE Conf. Comp. Vis. Patt. Recogn.}, 2014.

\end{thebibliography}
\end{document}